\documentclass[sigconf]{acmart}
\AtBeginDocument{%
  \providecommand\BibTeX{{%
    \normalfont B\kern-0.5em{\scshape i\kern-0.25em b}\kern-0.8em\TeX}}}

\renewcommand\footnotetextcopyrightpermission[1]{}
\setcopyright{acmcopyright}
\copyrightyear{2021}
\acmYear{2021}
\acmConference[ ]{ }{ }{ }
\usepackage{array}
\usepackage{tcolorbox}

\newcommand{\hush}[1]{}
\usepackage{graphicx}
\usepackage{subcaption}
\usepackage{mathtools}

\usepackage[LFE,LAE,T2A]{fontenc}
\usepackage[utf8]{inputenc}
\begin{document}

%\title{Fast-response Narrow-topic Text Classifiers: the COVID-19 Experience}
\title{Regular Expressions for Fast-response COVID-19 Text Classification}

\author{Igor L. Markov}
\email{imarkov@fb.com}
\affiliation{%
  \institution{Facebook Inc.,\\
  Menlo Park, CA}
}

\author{Jacqueline Liu}
\email{jacliu@fb.com}
\affiliation{%
  \institution{Facebook Inc., \\
  Menlo Park, CA}
}

\author{Adam Vagner}
\email{adamv@fb.com}
\affiliation{%
  \institution{Facebook Inc. \\
  Menlo Park, CA}
}

\renewcommand{\shortauthors}{Markov, Liu, and Vagner}

\begin{abstract}
 Text classifiers are at the core of many NLP applications and use a variety of algorithmic approaches and software.
 This paper introduces infrastructure and
 methodologies for text classifiers based on large-scale regular expressions.
 In particular, we describe how Facebook
 determines if a given piece of text --- anything from a hashtag to a post --- belongs to a narrow topic such as COVID-19. To fully define a topic and evaluate classifier performance we employ human-guided iterations of keyword discovery, but do not require labeled data. For COVID-19, we build two sets of regular expressions: (1) for 66 languages, with 99\% precision and recall >50\%, (2) for the 11 most common languages, with precision >90\% and recall >90\%. Regular expressions enable low-latency queries from multiple platforms. \hush{PHP, Python, Java and SQL code} Response to challenges like COVID-19 is fast and so are revisions. Comparisons to a DNN classifier show explainable results, higher precision and recall, and less overfitting. Our learnings can be applied to other narrow-topic classifiers.
\end{abstract}


%%
%% Keywords. The author(s) should pick words that accurately describe
%% the work being presented. Separate the keywords with commas.

%%
%% This command processes the author and affiliation and title
%% information and builds the first part of the formatted document.
\maketitle
\pagestyle{plain}

\section{Introduction}
\label{sec:intro}

Most application surfaces supported by Facebook currently implement special treatments of material related to COVID-19 \cite{facebook2020covid}. For example, in Search users with relevant search queries are presented with authoritative information about COVID-19. The frequencies of search queries in each language help determine salient keywords and check for unintended matches. On the other hand, our search context imposes a very strict interpretation of relevance to COVID-19 --- only words and expressions that name the virus or the illness are accepted. For example, “corona symptoms” and “sars-covid-2” are considered relevant, but “quarantine jokes” are not.
This strict interpretation is broadened for applications in page/group recommendations, News Feed ranking~\cite{ranking2021fb}, various integrity treatments~\cite{halevy2020integrity} (with customization often desired). Misspellings, such as “korrana vairos” are within the scope under all interpretations, but “corona beer” and “Corona del Mar, CA” are not. Among non-dictionary words considered, we note the dismissive jargon “covidiots”, and the ayurvedic treatment “coronil”. 

\vspace{1mm}
\noindent
{\bf Content types} handled by our solutions often intermix:

\begin{itemize}
\item 
URLs and Instagram hashtags that often merge words
\item Account, page and group names as well as search queries that usually separate words, but don’t form full sentences
\item 
Entire posts and individual comments
\end{itemize}

Checking entire posts calls for more subtlety than short-text categories, due to a broader vocabulary and longer phrases. The ability to handle many content types well sets our work apart from much of prior art
\cite{collobert2011natural,kim2014convolutional,regex_golf,li2008regexlearn,wang2020revisiting,chen2020multimodal,locascio2016neural}.

\vspace{1mm}
\noindent
{\bf Multilingual support and obstacles to using NLP tools.}
The global reach of Facebook-family products mandates support for the world's many languages, from Amharic to Vietnamese, and their diversity brings additional challenges; examples include Unicode support, complications of right-to-left languages, and the lack of sufficient data to automatically train classifiers in all languages. Automatic language detection is challenged by mixed-language content and short examples in related languages (Russian, Ukrainian and Bulgarian).
While automatic translation is useful on a case-by-case basis, it often misses context and nuance. Some languages use multiple terms or spellings for “corona” and “coronavirus”. In Arabic, we found six word forms for “corona” as a description of the virus. But in Spanish, “corona” ("crown") also appears in other contexts, showing up in geographic and personal names, making it a risky keyword for COVID-19 relevance.
The use of many NLP tools \cite{jurafsky-martin-book} is limited in our application by our support for many languages. For example, robust lemmatization solutions are not available for many languages, and even for English they do not handle the large number of misspellings of “corona”, “covid”, etc.
The use of word embeddings and transfer learning for neural nets is somewhat handicapped by the many emerging words and their spelling variants.
Reasonable training sets are difficult to find (especially for the less popular languages), 
short text lacks context, hashtags and URLs require dedicated tokenization, etc. But even for the most common languages, our solutions provide higher precision and recall without external dependencies, thus reducing online latency and software complexity. As a matter of careful performance evaluation,
it is important to know how much performance we leave on the table compared to a well-designed deep learning classifier on English posts, where tokenization is not critical, text length provides sufficient context, enough training data can be collected, and checking results does not require language expertise. Fortunately, our colleagues built a competent deep-learning solution based on \cite{collobert2011natural,kim2014convolutional}, validated it by manual labeling, and improved it iteratively. Results of an empirical comparison back our approach.

\vspace{1mm}
\noindent
{\bf Urgent change requests.}
Classifier coverage must be increased whenever new names for the virus (“NCoV 2019”, “SARS-CoV-2”) and the illness (“COVID-19”) arise, and in many other cases. On August 17, 2020 the misinformation researcher Ben Decker warned that Dr. Judy Mikovits --- the author of the conspiracy-theory “Plandemic” video --- was going to release a second video on August 18, using the word “indoctorination”.
According to the media on August 19 \cite{plandemic2020theverge},
the leading social-network platforms succeeded in limiting the distribution of this video, unlike for the first "Plandemic" video in May.
For such special cases, our classifiers can be revised quickly without labeled data. As part of due diligence, we found additional keywords (“plannedemic”) and similar but unrelated keywords, excluded to avoid false positives. To support urgent changes, we use regular expressions (regexes).

\vspace{1mm}
\noindent
{\bf Skepticism about regular expressions} abounds in the industry, despite common use at small scale. The infrastructure for large regexes remains limited, while some practitioners lack the mastery of regexes and technical understanding of pros and cons. An important insight is that {\em matching} is significantly easier than {\em parsing}~\cite{popov2012regexpower}.
Contrary to a common misunderstanding, 
modern practical regular expressions (regexes) can match many {\em context-sensitive} character configurations ({\em formal languages}) by using {\em backreferences} and {\em zero-width lookaround assertions}~\cite{popov2012regexpower,campeanu2003formal}. A regex can match "virus" if preceded by "corona", or match "corona" but not "corona beer". With two regexes, we can match the set-difference of two formal languages, but context-free languages are not closed under set-difference. Regexes with backreferences are NP-complete.
In applications, many engineers focus on deep learning solutions, leaving low-hanging fruit for other methods (Section \ref{sec:DNN}). We explain how to harvest this low-hanging fruit, noting gaps in the literature on regex-building (Section \ref{sec:conclusions}).
In academia, regex methods are criticized for hand-optimization.  However, developing competent ML models also requires manual effort, although of a different kind. Our regex methodology is faster (per language) than building ML models because data collection is automated. For serving, regexes require much simpler infrastructure and resources.

\vspace{1mm}
\noindent
{\bf In the remainder of this paper}, we describe our infrastructure to store and access regexes from several application platforms. We then explain how we scale regexes in size and improve their precision. Several aspects of regex syntax are key to our work; those we illustrate by example and show the regex patterns we use. The coverage of high-precision Tier 1 regexes is rounded up by a discussion of validation and our development methodology. Tier 2 regexes increase recall for a moderate loss in precision. Here we use automated translation and develop keyword discovery methods. To ensure explainability, our regexes produce matched snippets useful for application-based customizations and fast evaluation of regex performance. We pay special attention to evaluation protocols and illustrate them by examples. We optimize regexes for timing performance and introduce structural regex analysis to identify
portions that have minimal impact on recall. Empirically, regex matching is very fast. We summarize lessons learned from developing COVID-19 regexes, outline how our infrastructure and techniques can be applied to other classification challenges, and compare our work to prior literature.

Text detection techniques we describe in this paper are complemented by many others in production solutions~\cite{ranking2021fb,halevy2020integrity}, including algorithms and manual content review. Our examples illustrate proposed techniques by simplified variants of production solutions.

\section{Basic infrastructure}
\label{sec:basic}

The initial effort on COVID-19 classification at Facebook collected several keywords in high-prevalence languages to form short regular expressions. These regexes were populated in a Unicode-compatible key-value storage directly readable from PHP, Python, Java and SQL code in multiple existing applications. Compared to other classifier infrastructures, regexes are (1) easy to use, (2) very inexpensive,
(3) provide low latency and high throughput. Software integration complexity is low, and regex updates can be distributed worldwide in under one hour. However, the risk of mistakes was initially high and prospects for scalability in size unclear.

Our initial mandate for COVID-19 regexes required synonymy to either the virus or the illness, with very high precision ($>95\%$, in practice usually $>99\%$). This kept regexes short and allowed us to quickly extend coverage to over 60 languages with help from native speakers only for particularly challenging languages, such as Arabic and Thai. The availability of these initial regexes supported accelerated development of downstream applications for many surfaces in Facebook-family products and helped put together training data for a DNN-based classifier developed later. However, recall was clearly limited by the “strict mandate” of specific keyword matching, by the size of these regexes, and by the initial focus on the keywords with the highest prevalence. In some cases, the recall of the initial regexes was doubled by adding overlooked keywords or word forms, as well as by careful generalization. However, the greatest improvements were attained by developing Tier 2 regexes that were not limited by size or by strict semantic requirements.

Tier 2 regexes match many more keywords and expressions to trade off the exceptionally high precision of Tier 1 regexes for significantly higher recall. In 2020, we supported Tier 1 regexes for 66 languages and Tier 2 regexes for 11 languages.

Given that terms and text in English often appear in posts written predominantly in other languages, it is common practice to concatenate the English regex to each individual language regex before matching, or apply the two regexes separately. We do not recommend concatenating more than several language regexes because this may lead to very large regexes and unexpected cross-lingual matches. To illustrate, an early Czech regex included the overly aggressive clause “\verb/(v[iyý]r).*([ck]o[rn].[nr][ao])/” for “virus.*corona” and several spelling variants unintentionally matched “environmental pest control” in English. In this particular case, the dot was replaced with a more restrictive character class.

\vspace{-3mm}
\subsection{Scaling up regex size and precision}
\label{sec:scalingup}

Using common regex syntax \cite{li2008regexlearn}, Sections \ref{sec:scalingup} and \ref{sec:structure} introduce our higher-level structures and illustrate them for the COVID-19 topic.

Improving the recall of COVID-19 regexes clearly required larger regexes. Our infrastructure comfortably supports individual regexes up to 30KB, and some of the applications can handle megabyte-sized regexes. However, regexes with more than 80 characters, especially in foreign languages, become hard to read and awkward to maintain. Scaling the regexes up required adding structure, including support for visual line breaks (not line breaks in text) and comments ignored during matching. Simply splitting a regex adds line break characters, which we don’t want to match.
To this end, different regex engines support line breaks, but we prefer not to rely on engine-dependent matching options or maintain different regex variants for applications in PHP, Python, Java and SQL. 
Therefore, we have developed a portable way to support such comments and line breaks, compatible with major regex engines.
But before explaining the details, we first introduce necessary syntax and solve a different problem --- matching “corona” not followed by “ beer” or “ry” (as in “coronary”). 
Recall that the {\tt ({\bf ?!}<regex>)} syntax expresses a negative zero-width lookahead assertion; this means the regex “\verb/corona(?!(beer|ry))/” matches “corona”, “coronavirus”, but not “coronabeer” and not “coronary”. 
To add an optional space before “beer”, use “\verb/corona(?!(\\s?beer|ry))/”. 
To add up to 3 optional non-word characters before “beer”, use “\verb/corona(?!(\\W{0,3}beer|ry))/”. 
Our regexes also exclude “coronation”, “Corona, Queens”, etc.
Since all text is lowercased before matching, we only use lowercase characters. 
Cross-platform compatibility for negative lookahead covers the popular PCRE regex engine, but not the C++ re2 library.

Negative lookahead assertions also help to embed inert text into a regex. Consider the regex "{\tt \bf (!?x)x}". 
It tries to match “x”, but precedes it by a negative lookahead assertion that insists that “x” does not appear next, creating a logical contradiction. 
In other words, this regex cannot match anything. And neither can this regex: “\verb/(!?x)xhello/” which appends the inert text “hello” to the contradiction.
It can be conjoined with a meaningful regex as an alternative: “\verb/foo|bar|(!?x)xhello/”. 
Here we match “foo” or “bar”, but nothing else. We embed version numbers 
(“\verb/(!?x)x_Version_1234|foo/”) and section names (“\verb/foo|(!?x)x_Section_III|bar/”). Adding a line break within inert text does not affect matching"

\begin{tcolorbox}
\tt
foo|{\bf (!?x)x}\\
|Section\_III|{\bf (!?x)x}\\
|bar
\end{tcolorbox}

\hush{
% As it turns out, negative lookback always requires fixed length
To illustrate how infrastructure can limit regex syntax, we turn to negative zero-width lookback assertions that disallow matches preceded by certain text. We found that our PHP environment does not fully support them. Fortunately, the need for such assertions is limited and they can often be simulated using non-capturing groups, supported by all relevant regex engines.}

Multiplatform support entails several nuances. Given that PHP requires delimiters on both sides of regexes as in “/foo|bar/” or the like, the stored regex must be extrapolated into a PHP string, which resolves backslashes. 
For example the \verb/\b/ (word boundary) turns into a non-printable ASCII character for “backspace”. To avoid this, we double all backslashes in stored regexes: “\verb/\\b/”. While Python does not require extrapolation, it is easy to handle “\verb/\\/” correctly.

\subsection{Advanced regex syntax and regex structure}
\label{sec:structure}

Our regexes consist of sections, each section contains unrelated clauses, separated by the | character. Each clause typically targets one word or a pair of words separated by some number of characters, e.g., “corona news”. In rare cases, a clause targets a longer expression, such as “Severe acute respiratory syndrome”. Individual words may allow multiple word forms and spelling variants, expressed using character classes or more general alternatives of subword fragments. For example, the fragment “\verb/[ck][ao0]?r[ao0]n[ao]/” captures 48 different spelling variants of “corona”, including “carona”, “cor0na” and “korona”. However, this great power comes with great responsibility - we need to exclude “Koronadal”, “corona beer” and other irrelevant common words and phrases. Among several techniques we use are negative zero-width look-ahead assertions that we discussed earlier, illustrated by the following one-line example:

\begin{tcolorbox}
\begin{verbatim}
([ck][ao0]?r[ao0]n[ao](?!(ry|do|[ct]tion|\\W{0,3}
beer|dal|\\W*queens|\\W{0,2}(ca|california)
\end{verbatim}
\end{tcolorbox}

Words like “virus” are too general (low-precision) to be used individually, even if we exclude the most common irrelevant phrases. Such {\em weak keywords} are assembled in two-word clauses with multiple loosely-related alternatives for individual words --- {\em bipartite clauses}. This structure uses sets of weak keywords and is powerful in practice as it captures $N*M$ word combinations using $N+M$ keywords. The toy example below illustrates such a bipartite structure, accounting for word order. 

\begin{tcolorbox}
\begin{verbatim}
([ck][ao0]?r[ao0]n[ao]|c[o0]vid).{0,80}?(v[ai]i?r[aou]s|flu)
|(v[ai]i?r[aou]s|flu).{0,80}?([ck][ao0]?r[ao0]n[ao]|c[o0]vid)
\end{verbatim}
\end{tcolorbox}
Here, N=50 and M=13, so 650*2 different keyword combinations are captured with arbitrary characters in between. In practice, we capture more combinations by orders of magnitude.

To allow arbitrary characters between words, we initially used “\verb/.*/” but found that far-away matching words introduce spurious matches.
Multiple variable-length wildcards in the same regex, e.g., “\verb/severe.*acute.*respiratory.*syndrome/”
slow down the matching process. Therefore, we limit the distance between matched words and replace “\verb/.*/” with “\verb/.{0,80}/”. Measured in characters,
this max distance is language-dependent, e.g., French is more verbose than English, which is more verbose than Hebrew. For individual clauses, tuning maximal distance balances precision and recall. 

Recall that regex matching is {\em greedy} by default, i.e., it tries to find {\em the longest possible} matching fragment. However, we prefer {\em the shortest} matches as they do not merge unrelated phrases, take less time to find, and provide more concise explanations. In regex lingo, we are replacing greedy matching with reluctant matching, e.g., by using “\verb/.{0,80}?/” instead of “\verb/.{0,80}/”.

Short words such as “cov” and “sars” run the risk of matching some unintended longer words, such as “covariance” and “tsars”. To prevent unintended matches, we surround short words with word-boundary symbols \verb/\b/, often with provisions for hashtags (\verb/#/) and digits. In some cases, we limit the number and type of characters between the beginning/end of the string (or the \verb/#/ character) and the given word. These safeguards are illustrated by the following one-line regex fragment.

\begin{tcolorbox}
\begin{verbatim}
#.{0,50}?c[o0]vi[dt](.?19)?
|(\\b|\\d|_|#)c[o0]vi[dt](\\b|\\d|_)
\end{verbatim}
\end{tcolorbox}

For right-to-left (RTL) languages, e.g., Arabic and Hebrew, editing regexes is awkward because ASCII characters are sequenced left-to-right and language-specific characters right-to-left. Some editors show inconsistent visual effects. We found helpful to put parentheses around each word in RTL languages.

\subsection{Regex validation and basic development}
\label{sec:valid}

Our infrastructure improvements support scaling regexes to large sizes, where the risk of mistakes calls for testing infrastructure. We now perform instantaneous checks before a regex change is submitted for review. These checks cover backslashes that should be doubled (in \verb/\\b/, \verb/\\d/, etc), require a valid regex, try known matching and nonmatching examples. We intercept several known bug types (empty alternatives in \verb/||/, \verb/(|/ and \verb/|)/ ) \hush{, enforce several style restrictions (\verb/\\/s instead of bare spaces)} and ban regex syntax which SQL and PHP interpret differently.

The basic infrastructure above is sufficient to support large regexes, but has a number of blind spots, such as checks for regex-engine compatibility, numerical estimation of precision and recall, etc. Our early methodology for checking regex modifications was based on interactive SQL queries. It was automated using Python-based Jupyter notebooks that
\begin{itemize}
\item test the candidate regex in Python,
\item launch SQL queries comparing the candidate regex to the latest production regex to estimate recall gain and aggregate most common “new” and “lost” matches, 
\item can be posted readonly, linked from a reviewed regex change and cloned for future code changes. 
\end{itemize}
Matched user search queries capture many keywords that also appear in other contexts. Being short, queries repeat and can be sorted by frequency, unlike posts which tend to be unique. Table \ref{tab:SQLdiff} illustrates new and lost matches triggered by a change to the English regex. Going through lists of 50 most frequent matches usually makes it clear if a regex works as intended and if it hits many false positives. Unacceptable changes are rejected, but otherwise the next iteration attempts to improve recall (by adding or generalizing clauses) or prevision (by removing or narrowing clauses, e.g., adding negative lookaround). We iterate on both search queries and posts.

\begin{table}[tb]
    \centering
    \begin{tabular}{|l|l|}
\hline
{\sc 2690 new matches} & {\sc 230 lost matches} \\
\hline
corona news & corona beer  \\
corona news 24 & koronadal news update \\
zoom news corona live & brigada news koronadal update \\
\#coronacrisis & convit19 \\
zoom news live corona & corona beer virus joke \\
korona news & corona beer virus \\
corona news bangladesh & koronadal updates \\
nepal corona news & corona beer virus funny \\
\#coronanews & sharp coronado hospital \\
corona news nepal & coronation hospital \\
$\ldots$ & $\ldots$ \\
\hline
    \end{tabular}
    \caption{\label{tab:SQLdiff}
    Evaluating a change to COVID-19 regexes.
    }
\vspace{-6mm}
\end{table}

\section{Keyword discovery and evaluation}
\label{sec:discovery}

Tier 2 regexes increase recall by 30-900\% compared to Tier 1 regexes by adding many more keywords, expressions and bipartite terms. While they allow some false positives for the sake of increasing recall, both precision and recall are routinely >90\%. Negative lookahead assertions and other techniques help maintain precision >95\% in many cases. Such improvements require a greater development effort and additional automation. Table
\ref{tab:tier2} shows recall gains of Tier 2 regexes over Tier 1 regexes for 11 languages supported as of September 2020. We estimate precision in all cases to be >90\%.

\begin{table}[tb]
    \centering
    \begin{tabular}{|c|l|c|}
    \hline
       & \sc Language & \sc Recall gain \\
    \hline
    AR & Arabic  & 56.7\%\\
    BN & Bengali & 93.5\%\\
    EN & English & 68.4\% \\
    ES & Spanish & 79.2\%\\ 
    HI & Hindi & 50.4\% \\
    ID & Indonesian & 37.3\% \\
    FR & French &  59.2\% \\
    PA & Punjabi & \bf 9.62x \\
    PT & Portuguese & \bf 1.07 \\
    RU & Russian & \bf 1.13x \\
    VI & Vietnamese & \bf 3.06x \\
    \hline
    \end{tabular}
    \caption{\label{tab:tier2}
    Recall gain for Tier 2 regexes in eleven languages. The untypically large recall gains for Punjabi and Vietnamese are due to the inclusion of several English words. 
    Not shown here, Tier 1 regexes cover 66 languages.
    }
    \vspace{-8mm}
\end{table}

\subsection{Automatic translation and its pitfalls}
\label{sec:translate}

During keyword discovery, we routinely rely on automatic translation to understand words and sentences, but the process requires some human expertise. While unable to process massive amounts of information, people provide insights into candidate solutions, formulate generalization and notice identify unexpected phenomena. For example, through human insight we learned that in Bengali (Bangla), inserting a space in the term for “corona” (splitting it into “coro” and “na”) produces the words “do not” and in rare cases this meaning can appear even without a space character. Additionally, swapping the words for "virus" and "corona" may change the meaning of the word for “corona” keyword to “don’t”.

We found subword issues particularly difficult in Arabic, where “corona” can be spelled in different ways, but combining some spelling variants leads to irrelevant matches. Germanic and Slavic languages have numerous word forms, some of them irregular, that must be taken into account by regexes to avoid lengthy enumerations. Word meanings and connotations are important when trading off precision for recall. For example, common Kazakh queries with the word for “ковид” (“covid”)
include “ковид белгілері” (“signs of covid”).
This phrase can be added as a two-word regex clause or used in a larger bipartite clause as a generalization and to produce more explicable matches. However, being a very general word,
“белгілері” 
(“signs”) 
cannot be used by itself, compared to “symptoms” which has a medical connotation.

Automatic translation is sufficient to support regex development for many languages when used with our keyword discovery tools and large searchable corpora. Languages with sophisticated grammar and word forms, such as Arabic, require native-speaker expertise because (a) automatic translation can be misleading, especially for short phrases, and (b) constructing regexes without language expertise can miss relevant word forms while running the risk of false positives due to unexpected substring matches. Any fluent speaker can sanity-check individual keywords and regex clauses, while having a working knowledge of regexes can help even more by flagging pitfalls, suggesting improvements, and expressing word stems and spelling variants to increase recall without undermining precision. Successful regex development is possible without language fluency (via automated translation), but a fluent speaker can accelerate work for difficult languages.

\subsection{Keyword discovery techniques}
\label{sec:keywords}

 When adding a new language, one starts with “corona” and “corona virus”, used even in languages with non-Latin scripts. For example, the word “corona” has low precision in Spanish where it means “crown”, but high precision in Portuguese (where “crown” translates as “coroa”). Where relevant, 
 translations of "corona" and "coronavirus" should be considered as well, noting that many language groups share words and spelling variants (Indonesian and Javanese, languages that use Cyrillics, etc). \hush{We start early exploration with user search queries on Facebook because they are short, focused on common keywords, and can be counted because they repeat. However, search queries are biased to current events, celebrities, etc.}

Additional keywords can be found among words that appear most commonly in matching search queries, as well as posts. To this end, posts use a broader lexicon than search queries.  To illustrate, we randomly sample a large number of French-language posts with “covid” or “coronavirus”. Table \ref{tab:french} shows the most common words in those posts. These words can be additionally categorized by Wikipedia-text frequencies.
Here we see the original two keywords and their variants (covid1, cov, ncov, \_covid, covid\_, \_covid\_, covit, cov0, covi1, cov19, c0v1), as well as many other relevant keywords (virus, pandemie, symptôm/symptômes, quarantaine, confinement), including multiple word forms (testing, tests, testez, testé, testés, testee, testées) and typos (testte). Some words are foreign (china, chinese). Some words are short and must be used with word boundaries (sars, cas) while others are fairly generic (cas, contre, mort, nouvelles, chine). There are also several apparent hashtags (coronafakevirus, ensemblecontrelacovid). For words that cannot be used individually, we find common bigrams. Even with a limited understanding of French, one can build a compact but comprehensive regex. We initially erred on the side of high precision and later gradually relaxed precision to significantly increase recall.

\begin{table}[tb]
    \centering
    \begin{tabular}{|c|c|c|c|}
    \hline
    covid  & coronavirus & cas & corona \\
    contre & covid1 & virus  & test  \\
    covi0  & sars & tests & sympt\^omes \\ 
    morts  & nouvelle & confinement & test\'e \\
    mort   & testées  & testes & wuhan \\
    infection & tester & cov & dernier \\
    teste & epidemie & c0vid & derni\`ere \\
    mortalité & derniers & pandemie & symptomes \\
    chine & d\'ec\'ed\'ees & sympt\^ome & test\'ee \\
    contrer & derni\`eres & stopcovid & covid\_ \\
    d\'ec\'ed\'e & d\'ec\'ed\'ee & pneumoni & infections \\
    derniere & contrepoints & testicules & covidinfos \\
    d\'ec\'ed\'es & morte & ncov & neumonies \\
    symptomatiques & mortes & testes & covids \\
    \_covid & testing & mortalite & pandemic \\
    covit & testez  & mortel & covi1 \\
    mortalite & symptomatique & covi0 & cov19 \\
    corana & symptome & notocovid & c0v1 \\
    sympt\^omatique & coronafakevirus & \_covid\_ & testee \\
    symptoms & decede & decedent & symptomatic \\
    testte & masquecovid & testant & quarantine \\
    neumonie & d\'ec\'eder & china & chinese \\
    sympt\^om & confinements & kovid & contrevenant \\
    \hline
    \end{tabular}
    \caption{\label{tab:french}
    Starting with just "covid" and "coronavirus",  our keyword discovery finds related French words.
    }
\vspace{-9mm}
\end{table}

While labeled data are not necessary in our approach, even noisy data enable refinements that focus on “false positives” and “false negatives” for a given regex (many of these turn out 
correct upon inspection). Initially, we used sampled posts from known coronavirus groups, as positively-labeled examples, along with posts that matched our earlier regexes. Incidentally, this collection was also used by our colleagues to train a supervised classifier. Subsequently, we used classifier decisions precomputed for large numbers of posts. We used several types of keyword discovery techniques:

\begin{itemize}
    \item Adjacent co-occurrence: e.g., starting with “corona”, we may find “news”, “symptoms”, etc. This technique is useful to find clarifying words for bipartite clauses, e.g., “halloween masks” vs “surgical masks”.
    \item Most frequent words in true positives: e.g., starting with “coronavirus” and “covid-19”, we may discover “doctor”, “hospital”, etc. When including these keywords in the regex, one should check the incremental gains, as some words only appear next to stronger keywords.
    \item Most frequent words in false positives can indicate erroneous keywords, unintended matches, overlapping topics, etc.  Keywords found this way can be used in negative lookahead assertions. When valid keywords appear among false positives, this indicates incorrect labels (see Section \ref{sec:DNN}).
    \item Most frequent words in false negatives: e.g., starting with “corona”, we find in a Russian corpus
    respective terms in Russian. However, this turns up many prepositions and modal verbs. Hence, we compare word frequencies in false negatives to those in all texts, then rank keywords by the ratio of frequencies --- we seek words more likely to appear in false negatives than in general texts. This technique turned up the French word “quarantaine” when the regex misspelled it as “quarantine”. For Bengali and Punjabi, the most common missing words were in English.
    \item Most frequent word bigrams and trigrams with words frequently in false negatives and false positives. Bigrams and trigrams give context, make automated translation more effective, help distinguishing different word meanings, and suggest bipartite clauses for regexes.
\end{itemize}

Here is a small sampler of bigrams found in false negatives for English at one point during the regex development:
\begin{tcolorbox}
'social distancing', 'in lockdown', 'you touch', 'but liquor', 'self isolation', 'not social', 'non essential', 'to thank', 'home please', 'gloves masks', 'safe everyone', 'doctors nurses'
\end{tcolorbox}

This suggests adding bigrams “social.distancing”, “self.isolation” and “gloves.masks” as regex clauses, while watching out for false positives this may introduce. These clauses can also be generalized to allow for more interstitial characters between words as well as multiple word forms, such as “distanc(e|ing)”. Notably, “self-isolation” brings many false positives, but can be included if used with additional weak keywords.
For French, Portuguese and other languages, it was important to distinguish masks related to COVID-19 from cosmetic and entertainment masks. The manual regex revision process (Section \ref{sec:valid}) sometimes uses transformations similar to those in \cite{li2008regexlearn}, but we do not automate what people do quite well, and instead focus on big-data aspects of the problem.

\subsection{Explainability and customization}
\label{sec:explain}

Unlike entire posts, short strings do not require laborious scanning to understand which parts matched and why. However, in order to deal with longer texts, we leverage the ability of regex matching functions to return a full list of matches. For each matched text, a downstream table contains a list of up to 30 of the matched snippets, sorted by increasing length. For example, a long text may produce the following array of matches: {\tt [ "covid1", "virus disease (covid", "health department corona" ]}. This helps confirming correct matches without reading the entire post. 

Tier 2 regexes are optimized for more explainable matches by trying to extend each match to full words or entire expressions, and including known keywords that clarify meaning. For example, we match “pandemi[ac]” in English (allowing the Latin form of the word often used by doctors). “Pandemi” is good enough, but not as clear to the reader. Along the same lines, we try to match “cases of coronavirus” rather than just “cases of corona”. On the other hand, for interstitials between keywords, we replace the default greedy regex matching with reluctant regex matching that uses as few words as possible. In practice, these adjustments provide concise but clear explanations, particularly valuable for long texts.

\begin{tcolorbox}
 {\sc Original text:}
 Facebook Data for Good has a number of tools and initiatives that can help organizations respond to the {\bf COVID-19} pandemic. Here’s how you can make use of these tools: ~~ 
 https://dataforgood.fb.com/docs/{\bf covid19}/
\end{tcolorbox}

\begin{tcolorbox}
 {\sc Extracted keywords:}
 {\bf COVID-19}, {\bf covid19}
\end{tcolorbox}

Downstream applications may assume different criteria for COVID-19 relevance. To accommodate multiple applications without modifying regexes, our Tier 2 regexes are inclusive and open to customization. We distinguish several types of customization:

\begin{itemize}
    \item Remove or normalize some text during preprocessing. For example, leaving only the first $k$ lines of text would require matches early on and emphasize the significance of the topic.
    In another example, one can remove “vaping epidemic” to avoid false positives with “epidemic”, if this word is matched standalone. Normalization can be illustrated by stemming and lemmatization for Slavic and Germanic languages that use numerous word forms. \hush{Yet, the Tier 2 regex for Russian handles word forms well.}
    \item  Apply the regex only to the first $k$ words (or some fraction) of a long text. This addresses a known issue with texts largely unrelated to COVID-19 that add “masks required” or “stay healthy during quarantine” at the end.
    \item  Check the matches and discount those that lead to common false/undesirable positives in a particular application, e.g., “quarantine jokes”.
    \item Use additional “negative” regexes to weed out false positives. We found posts on lockdowns during hurricanes, surgical masks used during surgery, non-COVID viral pneumonias, vaccines development for non-COVID infections, etc.
\end{itemize}

\subsection{Performance evaluation with "safe" regexes}

To evaluate a change in a regex,
our basic regex development methodology (see above) uses search traffic to find the most common “new” and “lost” matches. We also use posts to estimate recall gain. Given that 
posts use longer phrases and a larger vocabulary than search queries, we use a more comprehensive evaluation later.

When working with (possibly noisy) labeled data, we zoom in on classifier disagreements and evaluate them manually. For example, seeing 10 actual false positives out of 200 formal “false positives”, (pessimistically) supports a >95\% precision bound for our regexes.

Evaluating 200 examples can be laborious even for a native speaker, whereas we do not understand most languages involved without translation. \hush{Moreover, posts are normally truncated to the first $n$ characters in spreadsheets, so the matching snippet may not even appear in the default view.} To streamline evaluation, we extract a list of matched snippets, as illustrated in the example on explainability.
While most posts are unique, their matched keywords exhibit statistics similar to those in search queries. Instead of always checking individual examples like the one in Section \ref{sec:explain}, we create a “safe” (100\% precision) regular expression for matched fragments. This safe regex includes “coronavirus” and automatically confirms many examples is true positive matches. A handful of such keywords usually confirm $>80\%$ of matches. For the Bengali (Bangla) language, adding just two non-English keywords to the general English safe regex covers $> 96\%$ “false positive” matches where the regex disagrees with the DNN classifier. Thus, out of 200 examples, we only need to check 7 with human intervention using automated translations, and they all come out correct, confirming a near-100\% recall. This process is fast and automated translation is now accessible from spreadsheet applications in batch mode. With a safe regex for a particular language, one can quickly iterate changes to the main regex. A more traditional (and laborious) human-labeling process can validate the final result, as discussed in Section \ref{sec:DNN}.

\subsection{Matching-time evaluation and optimization}

\begin{table}[b]
\vspace{-2mm}
    \centering
    \begin{tabular}{|c|c|c|c|c|}
    \hline
      Tier & Regex & \multicolumn{3}{|c|}{Runtime in $ms$ for non-matching text of size} \\
      & size &  1K chars & 3K chars & 5K chars \\
    \hline
    1 & 1.71KB & 1.35 ± 28.7$\mu s$ & 4.11 ± 140$\mu s$ & 6.94 ± 28.3$\mu s$  \\
    2 & 4.87KB &
    7.3 ± 96.5$\mu s$ &
    13.9 ± 141$\mu s$ & 
    14.3 ± 325$\mu s$ \\
    \hline
    \end{tabular}
    \caption{\label{tab:timing}
     Matching runtimes for English regexes.
    }
\vspace{-4mm}
\end{table}

Regex size is an important parameter for both maintainability and matching time which can limit the bandwidth of the COVID-19 classifier. Downstream tables may take hours to perform matching on all rows. We time regexes in Python and SQL on a small set of long posts, including matching and non-matching examples. Non-matching posts usually take longer. When matches exist, returning all of them excludes early termination. When matches and explainability are unimportant, one can perform matching with early termination. Among common techniques for regex-matching acceleration, we found that avoiding avoiding multiple “\verb/.*/” per clause is most impactful (Section \ref{sec:structure}).

\begin{table}[tb]
    \centering
    \begin{tabular}{|c|c|c|c|}
    \hline
& posts & search & max \\
\hline
\verb/.{0,80}?/ & 47.496 & 27.68 & 47.496\\
\verb/[vw][ai]i?r[aou]s|infection/ &
27.097 & 17.222 & 27.097 \\
\verb/i[dt]/ & 16.720 & 25.18 & 25.18 \\
\verb/c[o0]vi[dt]/ &  16.720 & 25.178 &
25.178 \\
\verb/\b|\d|_|#/ & 16.605 & 23.848 & 23.848 \\
\verb/\b|\d|_/ & 16.605 & 23.846 &
23.846 \\
\verb/[ck][ao0]?r[ao0]n[ao]/ &
18.636 & 16.76 & 18.636 \\
\verb/[ck][ao0]r[ao0]n[ao]/ & 8.630 &
2.082 & 8.630 \\
\verb/?c[o0]v/ & 0.024 & 1.432 & 1.432 \\
\verb/#.{0,50}?c[o0]vi[dt]/ & 
0.018 & 1.242 & 1.242 \\
\verb/news/ & 1.119 & 0.128 & 1.119 \\
\verb/today/ &  0.889 & 0.01 & 0.889\\
crisis & 0.767 & 0.06 & 0.767 \\
update & 0.092 & 0.616 & 0.616 \\
health & 0.582 & 0.042 & 0.582 \\
\verb/[ck][o0][nr]?vid/ & 0.034 & 0.562
& 0.562 \\
\verb/19|cdc/ & 0.043 & 0.482 & 0.482 \\
test & 0.409 & 0.116 & 0.409 \\
\verb/^|\b/ & 0.348 & 0.102 & 0.348\\
fight & 0.339 & 0.276 & 0.339 \\
china & 0.320 & 0.092 & 0.320 \\
lockdown & 0.302 & 0.036 & 0.302 \\
\verb/quar[ae]ntin.?/ & 0.183 & 0.276 &
0.276 \\
pandemic &  0.261 & 0.052 & 0.261 \\
against &  0.251 & 0.046 & 0.251 \\
\verb/[ck][ao0]?r[ao0]n[o0a]/ & 0.223 &
0.024 & 0.223 \\
hospital & 0.219 & 0.014 & 0.219 \\
case & 0.203 & 0.072 & 0.203 \\
\verb/c[o0]v.?\d/ & 0.0316 & 0.13 &  0.13\\
death & 0.1241 & 0.056 & 0.1241\\
$\ldots$ & & & \\
    \hline
    \end{tabular}
    \caption{\label{tab:profiling}
    Structural profiling of a regular expression. The numbers shown are \% of content samples {\em not} matched when a given fragment is replaced with {\tt foobar123} in the regex.
    Results for Facebook posts and user-issued search queries are given in separate columns.
    }
\vspace{-9mm}
\end{table}

Table \ref{tab:timing} gives timing results for regex matching in Python using \verb/re.search()/. Runtime scales predictably with text size and regex size, well into kilobytes. Matching is fast because all data fits in L2 cache (multiple MiBs in modern servers) and no floating-point calculations are used. This can be compared to common ML models that represent each word by 200-300 floating-point values and evaluate those with respect to millions or billions of weights, incurring latency 10-100ms.

In order to identify and remove unnecessary structural components of regexes, we implemented a tool that breaks down a given regex into clauses and chunks, then iteratively removes each to understand its contribution to recall. Clauses and chunks are then ordered by decreasing gains; this clearly identifies components that are very important. The tail of the list includes not only the least important components, but also useful components that are somewhat redundant (often intentionally, to improve coverage and/or make matches clearer).

In Table \ref{tab:profiling}, keywords and regex chunks are ordered by the fraction of matches lost when a chunk is removed. For example, “\verb/c[o0]vi[dt]/” is one of the most useful chunks, but “pneumonia” is among the least useful because “pneumonia” typically appears with other keywords, while “covid” doesn’t have direct synonyms or co-occurring words that are more powerful. Chunks that appear in multiple places rank near the top because we blank out all of their instances at once. This table was produced with full automation and is routinely rebuilt during various iterations of the regex.

To slim down a given regex and accelerate matching with minimal impact on recall, one can remove the chunks from the bottom of the table. However, there are pitfalls in this process.

\begin{itemize}
    \item As a thought experiment, create two copies of some very useful keyword (\verb/c[o0]vi[dt]|c[o0]vid/)
    and make a superficial modification in the second copy. As a result, each of the copies will move to the bottom of the list because removing one copy preserves most matches. This example shows that removing related keywords from the bottom of the list is risky. Therefore, we remove keywords in small groups only, if not one at a time.
    \item Keywords that are inessential for recall may be important to maintaining high precision when they rule out false matches.
    \item  Keywords that appear statistically inessential may catch rare but important spelling variants.
\hush{for example "c0r0na" is rare but more common in conspiracy theory materials.}
\end{itemize}

We term this process {\em regex distillation} and apply it when new clauses are added without estimating prevalence. In practice, distillation has a limited impact on smaller, more conservative regexes.

\section{Empirical comparison to a DNN}
\label{sec:DNN}

We have introduced a detailed methodology for developing, evaluating and incrementally improving COVID-19 regexes in multiple languages. Empirically, it produces compelling results used by a number of applications at Facebook with closely tracked products metrics. Our COVID-19 regexes have been supporting a variety of product surfaces and applications since March 2020.

The methodology has been successfully replicated for a variety of applications, many not related to COVID text matching, and faces no competition because it requires very modest resources and no training. For completeness, we compare our solution to a deep-learning classifier
based on modern pre-trained 1024-dimensional word embeddings and a CNN architecture from
\cite{collobert2011natural,kim2014convolutional}. It uses 100 kernels and
a 128-dimensional final dense (MLP) layer. Developed for Facebook posts in several languages, this classifier did well when evaluated by manual labeling and was iteratively improved over several weeks. More recent NLP architectures exist, but the issues seen in our comparisons are common to supervised classifiers.

Our evaluation methodology consists of sampling and manual labeling by people independent of the regex development process. This comparison was performed jointly for regexes and a DNN classifier to produce performance metrics for both on the same dataset. To ensure fairness, the regexes and the DNN classifier were finalized within three weeks from each other. DNN training data included COVID-related posts found by regex matching. Test data includes random samples and posts labeled positive by at least one of the two classifiers (this improves the accuracy of recall estimates) and ask a person to label them. The regex matches, when available, expose the matched fragments that serve as explanations and simplify human labeling. \hush{For example, seeing “coronavirus” or “covid symptoms” among matched fragments makes it unnecessary to read a long post.} As a side effect, the same labeling effort produces precision and recall estimates for both classifiers. This approach cannot be used in the regex development process because it is slow, but it can validate the final results.

First, we examine the “false positives” for the Tier 2 regex relative to the DNN classifier, i.e., posts that were labeled as COVID-related by the regex and not COVID-related by the DNN classifier. The most frequent words found in these posts are: covid, healthy, hospitalized, covid\_, coronovirus, updated, chinatown, pandemics, corona, chinese, fightscorona, convid, virus, death, newsweek, indiafightscorona, news, outbreak, treatments, corvid, health, testing, newsom, covid19, quarantine, quarantined, carona, cases, tests, and many more. From the 100 most frequent words, it is clear that the DNN classifier is missing a lot of relevant material, including direct keywords “covid” and “coronavirus”, their numerous spelling variants (covid\_, \_covid, covid\_19, convid, corvid, convid19, covid1, coronovirus, carona) as well as alternative names (sarscov, ncov) and emerging words like (ccpvirus). Numerous high-precision keywords (quarantine, pandemics) appear in this list, as well as hashtags (beatcovid19, fightcovid19, stopcovid19, indiafightscorona). The list is full of weak keywords that appear in COVID-related documents: virus/viruses, infection/infections, symptom/symptoms, case/cases, hospitalizations, death, crisis, test/testing, news, latest. Rather than enumerate all spelling variants and keyword combinations, our regexes capture them explicitly with character classes, word and subword alternatives, as well as bipartite clauses.

We inspect not only “false positives”, but also “false negatives”, i.e. posts flagged as COVID-related by the DNN classifier but not the regex. For example, for posts in Russian, none of the 20 most common "false negative" bigrams are COVID-related. Several of them are related to the August 2020 protests against rigged elections in Belarus. The trend to label anti-police and anti-government protests shows up in other languages as well, as we discuss in the context of English-language comparisons below.

A head-to-head comparison of COVID-19 regexes to the DNN classifier, \hush{(version 3, April 2020)} facilitated by manual labeling, serves several purposes:

\begin{enumerate}
\vspace{-1mm}
    \item Sample with both regex and DNN classifiers to find more positives and improve the accuracy of recall estimates.
    \item
Reuse human-labeling effort to evaluate multiple classifiers.
\item
Check if each of the given classifiers is viable.
\item
Check which classifier is more often correct.
\item
Estimate precision and recall for each classifier.
\item
Validate our evaluation methodology based on “safe regexes”.
\item
Find weaknesses in each classifier vs. the other classifier.
\vspace{-1mm}
\end{enumerate}

To comment on specific items:
{\bf (1)} Sampling with multiple classifiers can leverage a weaker classifier to sharpen recall estimates for stronger classifiers. {\bf (2)} The reuse of human effort during labeling is greater than the obvious 2x because matched fragments provided by regexes simplify the labeling process. In comparison, evaluating the DNN classifier alone would be much more laborious. {\bf (3)} Cursory comparisons of regex performance to DNN performance on 11 supported languages suggested that the DNN classifier was not viable for Bengali and Punjabi. Indeed, these languages were not supported by the DNN classifier, but simply “spilled through” automated language detection. This was easy to fix.

Items 4-6 are related in that our methodology with “safe regexes” also provides evaluations in items 4 and 5. To this end, the detailed evaluation performed for English validated our methodology, making detailed evaluation less critical for other languages. Weaknesses found in the the DNN classifier are discussed below.

In Table \ref{tab:DNN} we compare the performance of the DNN classifier to Tier 1 and Tier 2 regexes. The table reports precision, recall and accuracy, as well as the recall gain of the Tier 2 regex over the DNN classifier. Accuracy is high because most posts are obviously not COVID-related. As intended, the Tier 1 regex has a near-100\% precision, but recall is a modest 55\%. In contrast, the Tier 2 regex exhibits recall and precision in the 93-95\% range --- a very attractive tradeoff. The DNN classifier shows recall half-way between those of Tier 1 and Tier 2 regexes, but its precision is only 53\%. 
%We found many English keywords missed by the DNN classifier.

\begin{table}[tb]
    \centering
    \begin{tabular}{|c|c|c|c|c|}
    \hline
     & DNN & \multicolumn{2}{c|}{Regex} &
       Gain: T2 vs DNN \\
     &   & Tier 1 & Tier 2 &  \\
    \hline
    Accuracy  & 96.25\% & 98.18\% & 99.54\% & 3.42\% \\
    Recall    & 71.36\% & 55.56\% & 93.58\% & 31.14\% \\
    Precision & 52.83\% & 99.56\% & 94.99\% & 79.79\% \\
    \hline 
    \end{tabular}
    \caption{
    \label{tab:DNN}
    Empirical comparisons between regex-based and DNN-based COVID-19 classifiers on English-language posts with respect to human evaluation.
    }
\vspace{-8mm}
\end{table}

Table \ref{tab:mislabel} shows examples mislabeled by the DNN classifier, which  places too much trust on certain words, e.g., “medics” and “vaccine”. Additionally, the DNN classifier marks posts related to George Floyd protests as COVID-related, likely because it was trained on data from April 2020 when all protests were COVID-related.
Based on our analysis of the data, it appears that the DNN model overfit to April 2020 data centered around protests and lockdowns.
Manual evaluation of the DNN model in April 2020 showed stronger performance,
although that evaluation did not sample
positives with two different classifiers.
In Section \ref{sec:conclusions}, we mention several advantages regexes hold over DNNs in our context, but here we only note that human-constructed regexes offer more robust generalizations.

The manual-labeling comparison confirmed that our Tier 2 regexes attain much better precision and recall than the DNN classifier. However, the DNN classifier contributed to this result in two ways. First, the regexes were improved (before the manual comparison) through a process that focused on disagreements with the DNN classifier. Second, the DNN classifier was used to obtain a high-quality set of examples for the manual evaluation. The results match the precision estimates from our “safe regex” methodology for streamlined evaluation of regexes, giving further credence to the results for other languages. In particular, our methodology strongly suggests that regexes produce fewer false positives and fewer false negatives than the DNN classifier when such a comparison can be made. For all 11 languages supported by the DNN as of September 2020, we estimate regex precision above 90\%. Recall can be estimated by (a) computing gain over Tier 1 regexes, (b) by sampling “false negatives” vs the DNN classifier. For most languages supported by the DNN classifier and regexes (AR, EN, ES, FR, ID, PT, RU, VI) we estimate regex recall >90\%.

\begin{table}[b]
\vspace{-4mm}
    \centering
    \begin{tabular}{l}
        \hline
    \bf False positives \\
    \hline
    “the riots over the weekend, our thoughts”\\
“curfew end june 9th birmingham don't do shit else” \\
“a grown woman will isolate herself to protect her peace... \\it’s very personal" \\
“there was no autism in vietnam before
bill gates \\  brought his vaccines | dr [such and such], md" \\
    \hline
    \bf False negatives \\
\hline
    “doraemon (home quarantine) episodes 1 to 30” \\
“wear a mask or go to jail.
102 years ago during the spanish flu.” \\
“covid-19 is increasing the isolation of \\
some of the most vulnerable canadians.” \\
“suspect in custody after sucker-punching
dc police officer \\ who was attempting
to enforce social distancing rules” \\
“with all that is going on via the covid-19 pandemic we know \\ that is tough and money
is tighter now than ever[...]” \\
    \hline
    \end{tabular}
    \caption{Examples mislabeled by the DNN classifier.
    \label{tab:mislabel}
    }
\vspace{-6mm}
\end{table}

\section{Conclusions}
\label{sec:conclusions}

We have outlined the development of COVID-19 text classifiers in numerous human languages with a lightweight infrastructure and low-latency evaluation, based on semi-automated keyword discovery and performance evaluation with “safe regexes”. Development is fast and requires no labeled data, but can use (possibly noisy) labels when available. The iterative refinement of regexes can be viewed as on-demand "smart" sampling, and has been used in this capacity by our colleagues to train deep-learning classifiers.

Regex matching produces concise explanations via matched snippets and can be customized by pre- and post-processing. Our classifiers have been in production use for Facebook search queries, News Feed posts and comments \cite{ranking2021fb}, Facebook and Instagram Ads, Instagram hashtags, OCR and post text, as well as account, page and group names, etc. \hush{For posts, regex-based classification can be compared to a competing DNN solution, which regexes outperform in precision, recall, resource utilization and evaluation latency.} Regex matching integrates natively (w/o additional latency) into applications that use PHP, Python, Java and SQL, and can be revised in one hour. 
\hush{
Regexes can provide significant resource savings compared to neural-network solutions and more traditional ML techniques. In applicartions, where regexes are insufficient, they can offload some of the work or provide precision-improving features for more traditional ML classifier architectures.}

\vspace{1mm}
\noindent
{\bf Other suitable applications.}
The regex approach works particularly well when a small set of keywords and phrases capture the meaning as distinctively as they do in COVID-19 classification. Moreover, small sets of keywords are not a limitation because 
\begin{itemize}
\vspace{-1mm}
    \item regexes can capture combinatorially large sets of word forms and phrases in limited space, as illustrated in our work,
    \item our infrastructure and tools can handle large regexes.
\end{itemize}

Combinatorial regex structures such as {\em bipartite clauses} offer powerful generalization that capture exponentially many keyword configurations. Our approach lacks floating-point weights and fine tuning for decision boundaries, but includes tunable integer-valued parameters, such as maximal distances between specific keyword pairs, keywords and line ends. More parameters can be used in post-processing, e.g. via logistic regression on matched fragments. While regex-based structures are likely too limited for sophisticated NLP tasks, we have significantly extended their performance from what is commonly assumed. They can also be used with existing NLP tools as filters, samplers, features, etc.

\vspace{1mm}
\noindent
{\bf Comparisons with DNN-based classifiers.}
A carefully designed and incrementally improved DNN-based supervised-learning classifier for COVID-19 texts based on \cite{collobert2011natural,kim2014convolutional} 
lagged behind regex classifiers in our comparisons. More labeled data, better labeled data, and more frequent retraining can help, but getting high-quality labeled data in many languages is problematic. The need for regular refresh of training data may become a showstopper in practice. The problem at hand requires a mix of word-level and subword-level considerations, which plays to the strengths of regexes. \hush{Developing appropriate word-embedding techniques would be interesting, but such an approach seems much more complex and resource-intensive than our regex-based solutions. Indeed, our regexes fully fit into the CPU cache (L2) and do not require floating-point operations. This gives regexes significant advantages in low latency, low hardware requirements and no need for large precomputed datasets (embeddings, weights, etc).} Regexes also provide concise explanations of matches, are amenable to fast-response updates, and are easy to customize via post-processing. These tasks are difficult in traditional supervised learning.

\vspace{1mm}
\noindent
{\bf Prior work} explored building compact regular expressions for short-strings --- phone numbers, IP addresses, etc
\cite{li2008regexlearn,wang2020revisiting}. 
Their scalability is limited by example-driven state- and regex-induction methods. Peter Norvig's "regex golf" \cite{regex_golf} and subsequent efforts
size-optimize regexes on small and medium-sized testsets. Compared to such supervised learning, we do not assume training data, but can optionally use labeled examples with noisy labels. To evaluate regex performance, our methodology uses "safe regexes".

Prior papers emphasize automated regex synthesis for a given spec. This often limits the character sets (to binary, ASCII, etc) and the regex syntax used.
\hush{and sometimes justifies unusual regex operators.}
Without such constraints, we use advanced features common in deployed regex engines for Python, PHP, Java and SQL. We ensure that matching decisions can be explained, and we keep regexes human-readable so that they can be quickly modified upon request. By supporting large regexes, we cover more complex topics than phone numbers and ZIP codes.

In another line of research, the authors of \cite{locascio2016neural} train a neural network on a set of regular expressions and their natural-language descriptions, such as 
'Stars with a number and contains the word "dog"'.
In our context, a human can easily convert this type of desciption into regexes. Instead of automating the work that humans can do, we automate work that is difficult for people, such as searching large databases and tallying the results.
Several recent publications \cite{chen2020multimodal,ye2020sketchdriven} 
take labeled examples along with natural-language specifications, such as this spec from \cite{ye2020sketchdriven}: "The max number of digits before comma is 15 then accept at max 3 numbers after the comma."
Since we consider such tasks straightforward for people, we work with more conceptual specs. Multiple iterations of regex refinement help us achieve precision and recall above 90\% in important cases, whereas recent publications settle for lower precision to ensure full automation. With a human in the loop, we produce compelling solutions for each natural language considered in one to two days by automating the big-data aspects of regex development. In this methodology, the human mind provides out-of-the-box data insights, generalizations, and
critical analyses, and drives investigations. Resulting regexes support maintenance and revision, while producing concise explanations for their matches. 

\vspace{1mm}
\noindent
{\bf Acknowledgments}
We thank Walid Taha for suggestions on improving the paper.

%%
%% The next two lines define the bibliography style to be used, and
%% the bibliography file.
\vspace{-2mm}
\bibliographystyle{ACM-Reference-Format}
\bibliography{kdd2021}

%%
%% If your work has an appendix, this is the place to put it.
%\appendix

\end{document}